\documentclass[runningheads]{llncs}
\usepackage{amsmath}
\usepackage[T1]{fontenc}
\usepackage{graphicx}
\usepackage{tikz}
\usetikzlibrary{shadows}
\usetikzlibrary{positioning}
\usetikzlibrary{arrows.meta}
\usetikzlibrary{calc}
\usetikzlibrary{shapes.geometric}
\makeatletter
\newcommand\blfootnote[1]{%
  \begingroup
  \renewcommand\thefootnote{}\footnote{#1}%
  \addtocounter{footnote}{-1}%
  \endgroup
}
\makeatother

\begin{document}
\title{Open-Set Vein Biometric Recognition with Deep Metric Learning}
\titlerunning{Open-Set Vein Biometric Recognition}

\author {Paweł Pilarek\inst{1}\orcidID{0009-0003-5955-9909}\textsuperscript{*} \and \\
Marcel Musiałek\inst{1}\orcidID{0009-0009-4964-0547}\textsuperscript{*} \and \\
Anna Górska\inst{1}\orcidID{0000-0003-3305-8711}}
\authorrunning{P. Pilarek, M. Musiałek,  A. Górska}
\institute{$^1$Wroclaw University of Science and Technology, Wroclaw, Poland\\
\email{278796@student.pwr.edu.pl, 279704@student.pwr.edu.pl,  anna.gorska@pwr.edu.pl}}
\maketitle

\blfootnote{* Equal contribution}

\begin{abstract}
Most state-of-the-art vein recognition methods rely on closed-set classification, which inherently limits their scalability and prevents the adaptive enrollment of new users without complete model retraining. We rigorously evaluate the computational boundaries of Deep Metric Learning (DML) under strict open-set constraints. Unlike standard closed-set approaches, we analyze the impact of data scarcity and domain shift on recognition performance. Our approach learns discriminative \(\ell_2\)-normalised embeddings and employs prototype-based matching with a calibrated similarity threshold to effectively distinguish between enrolled users and unseen impostors. We evaluate the framework under a strict subject-disjoint protocol across four diverse datasets covering finger, wrist, and dorsal hand veins (MMCBNU\_6000, UTFVP, FYO, and a dorsal hand-vein dataset). On the large-scale MMCBNU\_6000 benchmark, our best model (ResNet50-CBAM) achieves an OSCR of 0.9945, AUROC of 0.9974, and EER of 1.57\%, maintaining high identification accuracy (99.6\% Rank-1) while robustly rejecting unknown subjects. Cross-dataset experiments evaluate the framework's generalisation across different acquisition setups, confirming that while the model handles large-scale data robustly, performance remains sensitive to domain shifts in low-data regimes. Ablation studies demonstrate that triplet-based objectives combined with a simple 1-NN classifier offer an optimal trade-off between accuracy and efficiency, enabling real-time deployment on commodity hardware.

\keywords{Vein Biometrics \and Open-Set Recognition \and Deep Metric Learning \and Domain Shift Analysis \and Computational Biometrics.}
\end{abstract}

\section{Introduction}
Biometric recognition has become a cornerstone of modern access control, offering higher security than password-based mechanisms. While adopted modalities such as fingerprint, iris, and voice recognition established significant benchmarks \cite{Jain2016_50Years,Singh2021_DL_Bio_Survey}, they face limitations: surface-level traits are susceptible to abrasion and forgery, while face and voice struggle with environmental noise and presentation attacks.

In this context, vascular (vein) biometrics offer a robust alternative due to high spoofing resistance. Located beneath the skin and captured via near-infrared (NIR) imaging, vein patterns are difficult to replicate and leave no latent traces \cite{FBI_VascularPattern,AlKhafaji2022_VeinReview}. Furthermore, vascular structures are considered stable over time and exhibit high distinctiveness.

Vein biometrics span finger, palm, dorsal, and wrist modalities. While early methods used texture descriptors, deep learning has recently become the dominant paradigm \cite{Hemis2025_HandVeinDL,Kocakulak2025_FingerVeinDL}, with convolutional neural networks (CNNs) achieving substantial gains in segmentation and matching. However, the vast majority of these methods are evaluated under a closed-set assumption \cite{Radzi2016_FV_CNN,Zhang2022_AGCNN_Vein}. In this scenario, all test identities are known beforehand, which fails to reflect realistic deployments where unseen subjects appear. A reliable system must not only recognize enrolled users but also reject non-enrolled (unknown) identities.

This aligns with open-set recognition, where models handle unknown classes absent during training \cite{Geng2018_OpenSetSurvey,Mahdavi2021_OpenSetSurvey}. While extensively studied in generic vision using strategies like calibrated scores or margin-based losses, open-set capabilities remain critical for security against unknown adversaries \cite{Abdullahi2023_OpenSetFusion,Li2025_OpenSetSurvey}. In contrast, open-set recognition for vein biometrics remains underexplored, with few works investigating specific architectures or loss functions \cite{Chen2021_ArcLossFingerVein,Abdullahi2023_OpenSetFusion}. This leaves key questions: how well can deep metric learning support open-set scenarios, and how should decision rules be optimized for effective rejection?

In this work, we address these questions by investigating open-set vein recognition within a deep metric learning framework. We learn an embedding function mapping subjects to a feature space and define a decision rule combining similarity scores with a calibrated rejection threshold. We further compare metric learning losses, analyzing their impact on the balance between recognition accuracy and rejection capability.

The proposed framework is applicable to various vein types (finger, wrist, dorsal) and is independent of specific acquisition setups. Validated on public datasets using realistic protocols, our experimental results (evaluated via EER and AUC) demonstrate that the learned embeddings maintain high recognition performance for enrolled users while providing effective rejection of unknown identities, indicating that deep metric learning is a promising direction for practical open-set vein biometric systems.

\section{Related Work}
\label{sec:related_work}

\subsection{Vein Biometrics and Computational Challenges}

Vein biometrics utilize internal vascular patterns captured via near-infrared (NIR) imaging, offering inherent resistance to spoofing and high privacy preservation \cite{FBI_VascularPattern,AlKhafaji2022_VeinReview}. While traditional hand-crafted pipelines based on texture or topology descriptors (e.g., maximum curvature, Gabor filters, LBP) established early benchmarks \cite{FBI_VascularPattern,AlKhafaji2022_VeinReview}, recent surveys indicate that their performance degrades significantly under strong intra-class variability and sensor changes. This computational limitation in modeling complex, non-linear variations necessitates the shift towards learned feature representations capable of handling open-set scenarios.

Concurrently, research has expanded into secure deployability. For instance, Ren et al. \cite{Ren2021_TemplateProtection_FV} proposed a finger-vein system processing RSA-encrypted images directly via a CNN. This underscores the necessity of balancing recognition accuracy with template protection—a key computational constraint in practical authentication systems which guides our proposed framework.

\subsection{Deep Learning for Vein Recognition}

Deep learning has reshaped the landscape of vein recognition over the last decade. Early work explicitly demonstrated that relatively shallow CNNs trained on finger-vein images can outperform traditional hand-crafted features \cite{Radzi2016_FV_CNN}. Radzi et al.\ reported high identification rates on small-scale datasets using a four-layer CNN, illustrating the potential of learned features even under limited data conditions. Recent surveys by Hemis et al.\ and Kocakulak et al.\ distinguish biometric methods based on the task—specifically classification versus verification—and the underlying neural architecture, covering classical CNNs, Residual Networks (ResNet) and Attention-based models \cite{Hemis2025_HandVeinDL,Kocakulak2025_FingerVeinDL}. These reviews conclude that such deep approaches consistently outperform traditional methods when combined with appropriate regularisation.

Several works have proposed specialised architectures tailored to vein patterns. Zhang et al. \cite{Zhang2022_AGCNN_Vein} introduced AGCNN, a network with learnable receptive fields and Gabor-like filters designed to capture line-like vein structures with reduced parameter complexity. Hybrid and multimodal systems have also been explored; for instance, Abdullahi et al. \cite{Abdullahi2023_OpenSetFusion} presented a network fusing fingerprint and finger-vein data using temporal modelling to exploit sequence information. Generative models have also found application in this domain, particularly for addressing data scarcity. Zhang et al. \cite{Zhang2019_GAN_Augmentation} utilized Generative Adversarial Networks (GANs) to synthesise realistic vein images, a strategy that supports training in data-limited or open-set scenarios.

Deep learning has also been extended beyond finger veins. Marattukalam et al. \cite{Marattukalam2023_WristVein_DL} proposed a lightweight deep learning system for contactless wrist vascular recognition, demonstrating efficient operation on low-cost hardware. Furthermore, the template-protection scheme by Ren et al. \cite{Ren2021_TemplateProtection_FV} illustrates how deep architectures can be integrated with cryptographic mechanisms (e.g., RSA-based encryption) to ensure security without compromising recognition accuracy.

Despite this progress, most deep learning–based vein recognition methods are still developed and evaluated under closed-set verification or identification protocols. Models are typically trained on a fixed set of subjects and tested on disjoint images of the same subjects, with performance reported in terms of standard error rates \cite{FBI_VascularPattern,Hemis2025_HandVeinDL}. Only a small number of works explicitly examine generalisation to unseen users \cite{Chen2021_ArcLossFingerVein}, and even fewer design decision rules specifically for open-set operation, leaving open the question of how well deep vein systems can distinguish between enrolled and non-enrolled identities.

\subsection{Open-Set Recognition in Biometrics}

Open-set recognition (OSR) addresses the problem of correctly classifying samples from known classes while explicitly handling samples from unknown classes that were not observed during training. Comprehensive surveys on OSR, out-of-distribution detection and open-world recognition categorise existing methods into discriminative, generative and hybrid approaches, and discuss evaluation protocols and applications across computer vision and pattern recognition \cite{Geng2018_OpenSetSurvey,Mahdavi2021_OpenSetSurvey,Li2025_OpenSetSurvey}. Common strategies include calibrating the outputs of closed-set classifiers, learning margin-based embeddings with compact intra-class clusters and large inter-class margins, modelling feature distributions and designing dedicated unknown detectors or rejection mechanisms.

In the biometric domain, OSR has been studied mainly for face recognition, but also for other modalities such as iris, gait and multimodal systems. Margin-based softmax variants and deep metric learning losses (e.g., ArcFace-style objectives) have been shown to improve open-set performance by enforcing better separation between genuine and impostor samples in the embedding space, often combined with feature-norm regularisation or score normalisation to stabilise decision thresholds \cite{Geng2018_OpenSetSurvey,Chen2021_ArcLossFingerVein}. More recent work proposes open-set biometric frameworks that jointly optimise closed-set recognition and unknown-user rejection, and evaluate models on protocols that more closely resemble operational conditions across multiple modalities \cite{Li2025_OpenSetSurvey}.

In contrast, open-set recognition for vein biometrics remains comparatively underexplored. Existing deep vein systems, including those based on CNNs, GAN-based augmentation, specialised architectures such as AGCNN, multimodal fusion networks and wrist-vein models, are typically assessed on closed-set protocols and do not explicitly consider unknown-user rejection \cite{Radzi2016_FV_CNN,Zhang2019_GAN_Augmentation,Zhang2022_AGCNN_Vein,Marattukalam2023_WristVein_DL,Abdullahi2023_OpenSetFusion,Ren2021_TemplateProtection_FV}. Although margin-based losses and deep metric learning have established strong open-set baselines in other modalities such as face and iris recognition \cite{Geng2018_OpenSetSurvey}, to our knowledge, there is no systematic study that evaluates these frameworks for vein recognition under explicitly open-set protocols that separate training, enrolment, and evaluation identities across multiple vein types (e.g., finger, wrist, dorsal). This gap motivates the present work, in which we bring together advances in deep metric learning and open-set recognition and investigate how embedding design, loss functions, and calibrated decision rules affect open-set performance in vein-based biometric systems.

\section{Proposed Method}
\label{sec:method}

In this section, we describe our proposed framework for open-set vein recognition. The system is designed to learn discriminative vascular representations from a set of known identities and subsequently use these representations to identify enrolled users while rejecting unknown subjects. The pipeline consists of three main stages: (1) data partitioning and preprocessing, (2) deep metric learning for embedding extraction, and (3) open-set enrollment and inference.

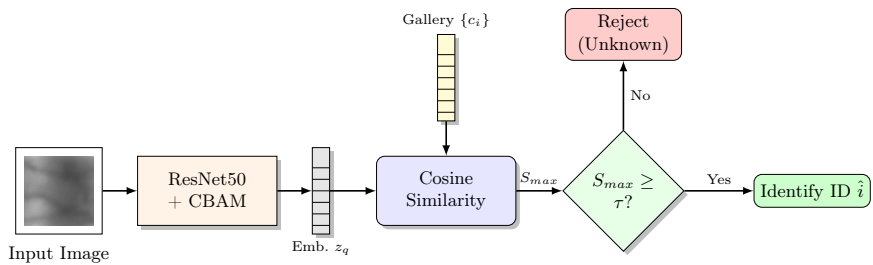
\begin{figure}[h]
\centering
\resizebox{0.96\textwidth}{!}{
    \begin{tikzpicture}[node distance=1.0cm, auto,
        block/.style={rectangle, draw, fill=blue!10, text width=2.2cm, text centered, rounded corners, minimum height=1.2cm, drop shadow},
        process/.style={rectangle, draw, fill=orange!10, text width=2.2cm, text centered, minimum height=1.2cm, drop shadow},
        decision/.style={diamond, draw, fill=green!10, text width=1.5cm, text badly centered, inner sep=0pt, drop shadow},
        vector/.style={rectangle, draw, fill=gray!20, minimum width=0.3cm, minimum height=1.5cm, drop shadow},
        line/.style={draw, -latex, thick}, 
        textlbl/.style={text width=2.5cm, text centered, font=\footnotesize}
    ]

        \node (input) [draw, fill=white, minimum width=1.5cm, minimum height=1.5cm] {\includegraphics[width=1.2cm, height=1.2cm]{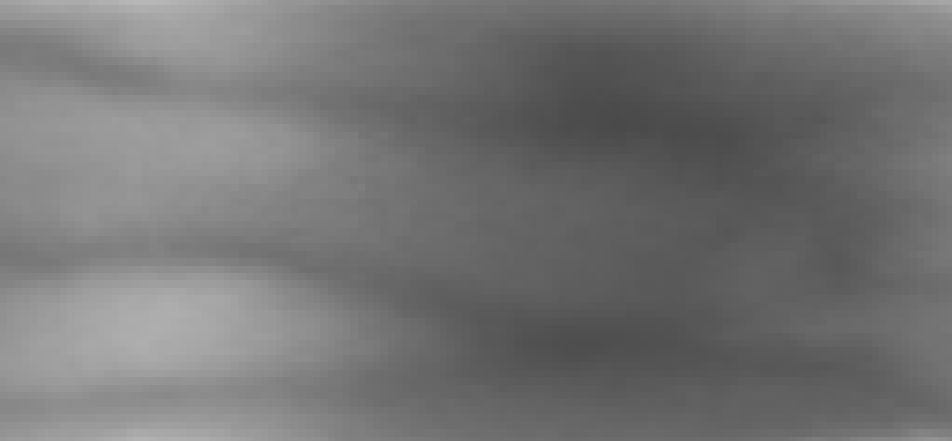}}; 
        \node[below=0.1cm of input] {Input Image};

        \node (backbone) [process, right=0.6cm of input] {ResNet50\\+ CBAM};

        \node (emb) [vector, right=0.6cm of backbone, label=below:{\scriptsize Emb. $z_q$}] {};
        \foreach \y in {-0.6,-0.4,...,0.6} \draw ($(emb.center)+(-0.15,\y)$) -- ($(emb.center)+(0.15,\y)$);

        \node (match) [block, right=0.8cm of emb] {Cosine Similarity};

        \node (gallery) [vector, above=0.6cm of match, fill=yellow!20, label=above:{\scriptsize Gallery $\{c_i\}$}] {};
        \foreach \y in {-0.6,-0.4,...,0.6} \draw ($(gallery.center)+(-0.15,\y)$) -- ($(gallery.center)+(0.15,\y)$);

        \node (dec) [decision, right=0.8cm of match] {$S_{max} \ge \tau$?};

\node (reject) [draw, fill=red!20, rounded corners, above=1.2cm of dec, text width=1.8cm, align=center] {Reject (Unknown)};
\node (accept) [draw, fill=green!20, rounded corners, right=1.2cm of dec, text width=1.8cm, align=center] {Identify ID $\hat{i}$};

        \path [line] (input) -- (backbone);
        \path [line] (backbone) -- (emb);
        \path [line] (emb) -- (match);
        \path [line] (gallery) -- (match);
        \path [line] (match) -- node[above, font=\scriptsize] {$S_{max}$} (dec);
        \path [line] (dec) -- node[above, font=\scriptsize] {Yes} (accept);
        \path [line] (dec) -- node[right, font=\scriptsize] {No} (reject);

    \end{tikzpicture}
}
\caption{Overview of the proposed open-set vein recognition framework. The input vein image is mapped to a compact, $\ell_2$-normalised embedding by the ResNet50-CBAM backbone. During inference, the system computes cosine similarity scores against enrolled class prototypes. If the maximum similarity $S_{max}$ falls below the calibrated threshold $\tau$, the sample is rejected as an unknown identity; otherwise, it is assigned to the nearest enrolled class.}
\label{fig:framework}
\end{figure}

\subsection{Open-Set Data Protocol and Preprocessing}

To simulate realistic open-set conditions, we enforce a strict subject-disjoint protocol. Given one or more public vein datasets (e.g., MMCBNU\_6000 and UTFVP for finger veins, FYO for wrist veins, and a dorsal hand-vein dataset), we treat each unique finger, wrist or hand instance as a distinct class. For each dataset, the subjects are first partitioned into two disjoint subsets: a known set used for training and enrolment, and an unknown set used exclusively to evaluate rejection capabilities. This ensures that the model never encounters the unknown identities during training.
Concretely, for MMCBNU\_6000 we use 60\% / 10\% / 30\% of the known identities for training, validation and testing, respectively. For UTFVP, the corresponding proportions are 50\% / 15\% / 35\%, while for FYO and the dorsal hand-vein dataset we use 55\% / 10\% / 35\% splits.
For evaluation, we adopt either a session-based or a sample-based protocol, depending on the dataset annotation. For datasets with explicit acquisition sessions, we use a session-based protocol in which enrolment images are taken from one session and probe images from different sessions. For datasets without session labels, we fall back to a sample-based protocol: for each enrolled class, a fixed number of images (e.g., 7 samples) are reserved for enrolment (gallery), while the remaining images (e.g., 3 samples) serve as probes for identification. When multiple datasets are used, we construct dataset-specific splits and optionally evaluate cross-dataset generalisation by using one database for training/enrolment and another as an open-set test source~\cite{FBI_VascularPattern,AlKhafaji2022_VeinReview}.

During training, we employ a balanced $P \times K$ sampling strategy to construct mini-batches, where each batch contains $P$ distinct identities and $K$ images per identity, providing sufficient positive and negative pairs for metric learning. Basic image pre-processing and data augmentation follow the experimental setup described in Section~4.2.

\subsection{Deep Embedding Framework}

\subsubsection{Backbone and Manifold Constraints.}
We utilize a ResNet50 backbone initialized on ImageNet and equipped with Convolutional Block Attention Modules (CBAM) to extract fine-grained vascular patterns. The network projects input ROIs into a compact embedding space ($D=512$). Crucially, we enforce an $\ell_2$ constraint on the output vectors, $||\mathbf{z}||_2 = 1$, restricting the feature manifold to a unit hypersphere. This normalization ensures that the optimization dynamics are driven purely by angular separability rather than magnitude variations inherent in NIR imaging, effectively stabilizing the training landscape.

\subsubsection{Metric Learning Objective.}
To structure this manifold for open-set discrimination, we employ the Batch-Hard Triplet Loss with a margin $m=0.3$. Unlike standard classification objectives, this loss operates directly on Euclidean distances. By utilizing a balanced $P \times K$ mini-batch sampler (where a \textit{class} corresponds to a specific anatomical instance, e.g., a unique finger), the objective dynamically mines the hardest positive and negative pairs. This explicitly minimizes intra-class variance while enforcing a fixed margin between disjoint identities, which is computationally essential for robust open-set rejection.

\subsubsection{Prototype-Based Inference.}
Following training, the system transitions to a prototype-based inference mode. For each enrolled identity $i$, we compute a class prototype $\mathbf{c}_i$ as the $\ell_2$-normalized centroid of its enrollment embeddings. This representation allows for efficient similarity-based matching and naturally supports incremental enrollment without the need for backbone retraining.

\subsection{Open-Set Decision Rule}

For open-set recognition, the system must decide whether a query sample belongs to one of the enrolled identities or should be rejected as unknown. Given a query image \(\mathbf{q}\), we compute its embedding \(\mathbf{z}_q = f(\mathbf{q})\) and the cosine similarity scores to all class prototypes:
\[
s_i = \mathbf{c}_i^\top \mathbf{z}_q, \quad i = 1,\dots,C,
\]
where \(C\) is the number of enrolled identities. Let
\[
S_{\max} = \max_i s_i, \quad \hat{i} = \arg\max_i s_i.
\]
The final prediction is given by a thresholded maximum-similarity rule:
\begin{equation}
\text{prediction} =
\begin{cases}
\hat{i}, & \text{if } S_{\max} \geq \tau, \\
\text{Reject (Unknown)}, & \text{if } S_{\max} < \tau,
\end{cases}
\end{equation}
where \(\tau\) is a decision threshold.

To determine an operating threshold \(\tau\), we use a validation set containing both known identities (disjoint from the enrollment set) and pseudo-unknowns (additional held-out known identities that are not enrolled). We sweep a range of candidate thresholds and select the value that maximises an open-set criterion (e.g., the Open-Set Classification Rate (OSCR) or AUROC) or satisfies a prescribed operating point such as a maximum tolerated false-positive rate on unknowns. This procedure is consistent with evaluation practices in the open-set recognition literature \cite{Geng2018_OpenSetSurvey,Li2025_OpenSetSurvey}.

\subsection{Evaluation Metrics}

We assess the system using both classical biometric metrics and open-set recognition metrics. For verification-oriented evaluation on the known subset, we report the Equal Error Rate (EER) and the True Positive Rate (TPR) at fixed False Positive Rate (FPR) operating points (e.g., 1\% and 0.1\% FPR), which are standard in vein biometrics \cite{FBI_VascularPattern,AlKhafaji2022_VeinReview}.

To quantify open-set performance, we treat the problem as a joint identification-and-detection task and report: (i) the Area Under the Receiver Operating Characteristic curve (AUROC) for distinguishing known versus unknown samples based on \(S_{\max}\), and (ii) the Open-Set Classification Rate (OSCR) curve \cite{Geng2018_OpenSetSurvey}. The OSCR curve plots the Correct Classification Rate (CCR) of known samples against the FPR on unknown samples as the decision threshold varies, providing a more informative view of the trade-off between accurate identification of enrolled users and rejection of impostors. Where appropriate, we summarise the curve either by its area (AUOSCR) or by CCR at specific FPR operating points, following recent open-set evaluation protocols\cite{Li2025_OpenSetSurvey}.

\section{Experiments}
\label{sec:experiments}

In this section, we detail the experimental setup, including the datasets, training configuration and evaluation protocols used to validate our open-set vein recognition framework. We then present quantitative results and ablation studies that analyse the impact of key components such as loss functions, embedding dimension and decision rules.

\subsection{Datasets and Splits}

We evaluate our method on four public vein biometric datasets: MMCBNU\_6000 and UTFVP for finger veins, FYO for wrist veins and a dorsal hand-vein dataset.
\begin{itemize}
  \item \textbf{MMCBNU\_6000 Finger Vein:} This dataset contains finger-vein images from 100 subjects, with six fingers per subject (index, middle and ring fingers of both hands), resulting in 600 unique classes. Each class typically has around 10 samples.
  \item \textbf{UTFVP Finger Vein:} A widely used finger-vein benchmark with 60 subjects and 360 classes, containing multiple images per finger captured under controlled near-infrared illumination. The dataset provides 1,440 images in total.
  \item \textbf{FYO Wrist Vein:} A contactless wrist-vein dataset acquired with an NIR system, where each wrist is treated as a separate class. The dataset comprises 160 subjects with 160 classes and 1,920 images.
  \item \textbf{Dorsal Hand Vein:} A dorsal hand-vein dataset with 251 subjects and 502 classes (left and right hands treated as distinct classes), containing 1,782 images.
\end{itemize}

Table~\ref{tab:datasets} summarises the basic statistics of the datasets used in our experiments, including the number of subjects, classes and images for each modality.

\begin{table}
\caption{Statistics of the vein datasets used in the experiments.}
\label{tab:datasets}
\centering
\begin{tabular}{ l c c c c }
\hline
Dataset~~       & ~~\#Subjects~~ & ~~\#Classes~~ & ~~\#Images~~ & ~~Modality          \\
\hline
MMCBNU\_6000  & 100        & 600       & 12,000      & Finger vein       \\
UTFVP         & 60         & 360       & 1,440       & Finger vein       \\
FYO           & 160        & 160       & 1,920       & Wrist vein        \\
Dorsal        & 251        & 502       & 1,782       & Dorsal hand vein  \\
\hline
\end{tabular}
\end{table}

To simulate a realistic open-set scenario, we enforce a strict subject-disjoint protocol. For each dataset, the total set of subjects is partitioned into two disjoint subsets: a \emph{known} set (used for training and enrollment) and an \emph{unknown} set (used exclusively for testing rejection). In our experiments, we typically use 70\% of subjects as known and 30\% as unknown. Within the known classes, we employ a session-based split: for each class, a fixed number of samples (e.g., 7 images) are reserved for enrollment (gallery), while the remaining samples (e.g., 3 images) are used as probes for identification. This ensures that the model is evaluated on unseen images of known subjects as well as completely unseen identities.

\subsection{Training Configuration}

The embedding network is trained using a balanced batch (\(P \times K\)) sampler. In each training iteration, we randomly sample \(P = 16\) identities and \(K = 4\) images per identity, creating a mini-batch of size \(P \cdot K = 64\). This strategy ensures that every batch contains sufficient positive and negative pairs for effective metric learning.

Input images are preprocessed by converting grayscale samples to three-channel RGB (to ensure compatibility with standard backbones), resizing them to a fixed resolution of \(224 \times 224\), and applying standard ImageNet normalization. To mitigate overfitting, we employ a streamlined data augmentation strategy during training consisting of random horizontal flipping (with probability \(p=0.3\)), and apply a dropout rate of \(0.5\) in the embedding head. We employ a ResNet50 backbone equipped with CBAM modules as our default architecture, natively producing \(512\)-dimensional \(\ell_2\)-normalised embeddings.

The model is optimized using the AdamW optimizer with a low initial learning rate of \(3 \times 10^{-5}\) (for stability under metric learning objectives) and an aggressive weight decay of \(10^{-3}\). A cosine-annealing scheduler is used to adjust the learning rate over 100 epochs. To prevent overfitting, we employ early stopping with a patience of 15 epochs, monitoring the validation loss on a held-out subset of the known classes (\(\Delta_{min} = 10^{-4}\)).

All experiments were conducted on a workstation equipped with a single NVIDIA GeForce RTX 4070 Ti SUPER GPU. The average training time for the ResNet50 backbone on the MMCBNU\_6000 dataset was approximately 30 minutes.

\subsection{Evaluation Protocol}

Our open-set evaluation protocol consists of three phases: enrollment, calibration, and testing.
\begin{enumerate}
  \item \textbf{Enrollment:} For each known class in the test split, we compute a class prototype by averaging the embeddings of the 7 enrollment samples. These prototypes form the gallery.
  \item \textbf{Calibration:} We use a separate validation set (containing both known and pseudo-unknown samples) to calibrate the open-set decision threshold. We sweep over candidate threshold values and select the one that maximises the Open-Set Classification Rate (OSCR) or AUROC, balancing the trade-off between correct classification of knowns and rejection of unknowns.
  \item \textbf{Testing:} During inference, query samples from both the known (probe) and unknown sets are compared against the enrolled prototypes. We employ a 1-Nearest Neighbour (1-NN) rule (or, in ablation studies, a \(k\)-NN variant) to determine the maximum similarity score. If the score exceeds the calibrated threshold, the sample is assigned to the corresponding class; otherwise, it is rejected as unknown.
\end{enumerate}

\subsection{Metrics}

We report a comprehensive set of metrics to assess performance:
\begin{itemize}
  \item \textbf{Detection Metrics:} AUROC (Area Under the ROC Curve) and EER (Equal Error Rate) measure the system's ability to distinguish between known and unknown subjects based on similarity scores. These metrics are widely established as standard benchmarks for open-set recognition in biometrics literature \cite{Geng2018_OpenSetSurvey,Chen2021_ArcLossFingerVein}.
  \item \textbf{Open-Set Metrics:} The OSCR curve plots the Correct Classification Rate (CCR) against the False Positive Rate (FPR). We also report the True Positive Rate (TPR) at strict FPR targets, e.g.\ TPR@FPR \(= 10^{-2}\) and \(10^{-3}\).
  \item \textbf{Identification Metrics:} For known subjects, we report Rank-1 and Rank-5 identification accuracy (CMC ranks).
  \item \textbf{Operational Metrics:} We report the final calibrated threshold, the accuracy on known subjects and the rejection rate on unknown subjects, as well as average runtime for enrollment and query inference.
\end{itemize}

All reported results are averaged over three random splits of the subjects into known and unknown sets, unless stated otherwise.

\subsection{Results and Ablation Studies}

\subsubsection{Overall Results and Architecture Comparison}
Comparison with State-of-the-Art: Direct comparison with existing vein recognition methods is challenging due to divergent evaluation protocols. While most prior works report results under closed-set assumptions (e.g., \cite{Radzi2016_FV_CNN,Zhang2022_AGCNN_Vein} reporting \textgreater{}99\% accuracy), our evaluation enforces a stricter open-set protocol with unseen impostors. However, our method's closed-set performance (Rank-1 \textgreater{}99.6\% on MMCBNU\_6000) is competitive with state-of-the-art closed-set systems, confirming that the open-set capability does not compromise identification accuracy for enrolled users.

We first compare different backbone architectures within our framework on the MMCBNU\_6000 dataset. Table~\ref{tab:arch} reports the main open-set metrics (OSCR, AUROC, EER) together with closed-set Rank-1 accuracy and average inference time per query. This comparison allows us to select the best-performing architecture for subsequent experiments on cross-dataset generalisation and ablation studies.

We selected the MMCBNU\_6000 dataset for backbone ablation and hyperparameter tuning primarily due to its scale and diversity. With 100 subjects, 600 distinct finger classes and 6,000 images, it provides a robust testbed for model selection.

\begin{table}
\centering
\caption{Architecture comparison on the MMCBNU\_6000 dataset. We report OSCR, AUROC, EER, Rank-1 accuracy and average inference time per query.}
\label{tab:arch}
\begin{tabular}{ l c c c c c c }
\hline
Model~~            & ~~OSCR~~   & ~~AUROC~~  & ~~EER~~    & ~~R1~~     & ~~Accuracy~~ & ~~Time {[}ms{]} \\
\hline
ResNet50-CBAM    & 0.9945 & 0.9974 & 1.5714  & 0.9960 & 0.9833   & 1.7685        \\
Attention U-Net  & 0.9894 & 0.9946 & 2.6825  & 0.9929 & 0.9627   & 2.4631        \\
CNN              & 0.9867 & 0.9927 & 2.8452  & 0.9921 & 0.9746   & 1.5907        \\
U-Net embedding  & 0.6314 & 0.7769 & 29.6071 & 0.7516 & 0.5317   & 2.6717        \\
\hline
\end{tabular}
\end{table}

Based on these results, we select the ResNet50-CBAM backbone as our default model in the remaining experiments. Unless stated otherwise, all subsequent results and ablations are obtained with this backbone.

\subsubsection{Cross-Dataset Generalisation and Domain Shift Analysis}

To assess the robustness of the selected model across acquisition setups, we evaluate it on all four datasets. Table~\ref{tab:generalization} summarises the cross-dataset performance. While the framework maintains competitive open-set recognition on MMCBNU\_6000 and FYO, we observe a marked performance drop on the UTFVP dataset (OSCR 0.76 vs 0.99 on MMCBNU).

\textbf{Computational Analysis of Data Scarcity:} This disparity reveals a critical dependency of Deep Metric Learning on the \textit{scale of the training manifold}. MMCBNU provides 12,000 images for manifold approximation, whereas UTFVP offers only 1,440. Our analysis suggests that in strict open-set protocols, deep models require a critical mass of diverse samples to learn acquisition-agnostic features. The high EER on UTFVP (31.54\%) indicates that without sufficient data density, the embedding space becomes sensitive to domain-specific artifacts (e.g., illumination variations), highlighting the necessity for transfer learning or synthetic data augmentation in small-scale medical scenarios.

\begin{table}
\centering
\caption{Cross-dataset performance of the proposed model on finger-vein (MMCBNU\_6000, UTFVP), wrist-vein (FYO) and dorsal hand-vein datasets.}
\label{tab:generalization}
\begin{tabular}{ l c c c c c c }
\hline
Dataset~~       & ~~OSCR~~   & ~~AUROC~~  & ~~EER~~    & ~~R1~~     & ~~Accuracy~~ & ~~Time {[}ms{]} \\
\hline
MMCBNU\_6000  & 0.9945 & 0.9974 & 1.5714  & 0.9960 & 0.9833   & 1.7685        \\
FYO           & 0.9746 & 0.9858 & 4.9424  & 0.9848 & 0.9627   & 2.4631        \\
Dorsal        & 0.9226 & 0.9251 & 13.3513 & 0.9963 & 0.9746   & 1.5907        \\
UTFVP         & 0.7576 & 0.7896 & 31.5418 & 0.9150 & 0.5317   & 2.6717        \\
\hline
\end{tabular}
\end{table}

\subsubsection{Ablation on Loss Functions}

We next investigate the impact of different loss functions on open-set performance. Table~\ref{tab:loss_ablation} reports the results for triplet loss, triplet loss with an additional centre regularisation term and contrastive loss, using the same backbone and training configuration. The comparison highlights which objective yields better separation between known and unknown score distributions.

\begin{table}
\centering
\caption{Ablation study on loss functions (MMCBNU\_6000 dataset, fixed backbone and embedding dimension).}
\label{tab:loss_ablation}
\begin{tabular}{ l c c c c c c }
\hline
Loss~~             & ~~OSCR~~   & ~~AUROC~~  & ~~EER~~    & ~~R1~~     & ~~Accuracy~~ & ~~Time {[}ms{]} \\
\hline
Triplet          & 0.9925 & 0.9956 & 1.8968  & 0.9952 & 0.9833   & 1.7685        \\
Triplet + Center & 0.9920 & 0.9955 & 1.7619  & 0.9944 & 0.9627   & 2.4631        \\
Contrastive      & 0.9770 & 0.9897 & 4.1865  & 0.9873 & 0.9746   & 1.5907        \\
\hline
\end{tabular}
\end{table}

\subsubsection{Ablation on Embedding Dimension}

Table~\ref{tab:dim_ablation} examines the effect of the embedding dimensionality on recognition and detection performance. We vary the size of the embedding vector (128, 256, 384, 512) while keeping all other factors fixed. The results reveal the trade-off between representation capacity and compactness.

\begin{table}
\centering
\caption{Ablation study on embedding dimension (MMCBNU\_6000 dataset, fixed backbone and loss).}
\label{tab:dim_ablation}
\begin{tabular}{ l c c c c c c }
\hline
Embedding dim~~ & ~~OSCR~~   & ~~AUROC~~  & ~~EER~~    & ~~R1~~     & ~~Accuracy~~ & ~~Time {[}ms{]} \\
\hline
128           & 0.9945 & 0.9974 & 1.5714 & 0.9960 & 0.9833   & 1.7685        \\
256           & 0.9920 & 0.9955 & 1.7619 & 0.9944 & 0.9746   & 1.5907        \\
384           & 0.9846 & 0.9930 & 3.2381 & 0.9889 & 0.5317   & 2.6717        \\
512           & 0.9924 & 0.9966 & 1.5992 & 0.9944 & 0.9627   & 2.4631        \\
\hline
\end{tabular}
\end{table}

\subsubsection{Ablation on the Decision Rule}

Finally, we analyse the influence of the decision rule by comparing the standard 1-NN classifier with \(k\)-NN variants that average the top-\(k\) prototype similarities. Table~\ref{tab:knn_ablation} shows that modest values of \(k\) do not improve open-set performance in our setting: while Rank-1 accuracy remains high, OSCR and AUROC degrade and EER increases as \(k\) grows.

\begin{table}
\centering
\caption{Impact of the number of neighbours \(k\) in the decision rule (MMCBNU\_6000 dataset).}
\label{tab:knn_ablation}
\begin{tabular}{ l c c c c }
\hline
\(k\)~~ & ~~OSCR~~   & ~~AUROC~~  & ~~EER~~     & ~~R1     \\
\hline
1     & 0.9920 & 0.9955 & 1.7619  & 0.9944 \\
3     & 0.8848 & 0.8889 & 18.8214 & 0.9937 \\
5     & 0.7949 & 0.7991 & 27.3571 & 0.9937 \\
\hline
\end{tabular}
\end{table}

\subsubsection{Practical Deployment Considerations}
From a deployment perspective, the proposed prototype-based framework offers significant scalability advantages. Memory requirements for the embedding gallery are minimal: each enrolled class is represented by a single $512$-dimensional vector (approximately 2 KB in 32-bit precision), meaning a database of one million identities requires only around 2 GB of memory. Furthermore, real-time enrolment of new users is seamless, requiring only a single forward pass to compute and store the new prototype without any network retraining. For highly scaled deployments, the exhaustive 1-NN search can be easily replaced with Approximate Nearest Neighbor (ANN) search algorithms to maintain real-time throughput.

\section{Conclusion}
\label{sec:conclusion}

We presented a computational framework for open-set vein biometric recognition, explicitly analyzing the boundaries of deep metric learning under strict subject-disjoint protocols. By constraining the embedding manifold to a unit hypersphere and optimizing it via batch-hard triplet loss, our ResNet50-CBAM model achieves highly competitive open-set performance on large-scale benchmarks (OSCR 99.45\% on MMCBNU\_6000).

Crucially, our cross-dataset evaluation reveals a significant computational trade-off: while the method demonstrates high robustness on large datasets, performance degradation on smaller subsets (e.g., UTFVP) highlights the sensitivity of metric learning to the density of the training manifold. This confirms that while deep embeddings provide a solid foundation for secure biometrics, their deployment in data-scarce medical environments requires careful consideration of domain shift and sample diversity.

Future work will focus on addressing these computational constraints. First, to specifically mitigate the sensitivity to domain shift and data scarcity observed in our cross-dataset evaluations, we aim to investigate domain adaptation and generative synthetic data augmentation techniques. Second, we plan to explore uncertainty estimation methods to refine rejection thresholds dynamically. Third, given the privacy-sensitive nature of biometric data, we intend to integrate template protection mechanisms (e.g., cancelable biometrics) directly into the embedding pipeline. Finally, we will extend our evaluation to continual learning scenarios, enabling the system to adapt to new subjects without full retraining.

%
%
%
\bibliographystyle{splncs04}
\bibliography{bibliography}
\nocite{MMCBNU_6000, Ton2013UTFVP, Toygar2020FYO, Lu2013RobustFV}

\end{document}